\documentclass{article}
\usepackage{spconf,amsmath,graphicx}
\usepackage{soul,color, cite}
\usepackage{amsmath,amssymb,amsfonts}
\usepackage[ruled]{algorithm2e}
\usepackage{graphicx}
\usepackage{textcomp}
\usepackage{xcolor}
\usepackage{subcaption}
\usepackage{hyperref}

\def\BibTeX{{\rm B\kern-.05em{\sc i\kern-.025em b}\kern-.08em
    T\kern-.1667em\lower.7ex\hbox{E}\kern-.125emX}}

\title{Multi-step Online Unsupervised Domain Adaptation}
\name{J. H. Moon, Debasmit Das and C.S. George Lee
}
\address{Purdue University, West Lafayette, IN, USA}
\begin{document}
\pagestyle{plain}
\ninept
\maketitle

%
%
%
%
\pagestyle{plain}
\maketitle

\begin{abstract}
In this paper, we address the Online Unsupervised Domain Adaptation 
(OUDA) problem,
where the target data are unlabelled and arriving sequentially.
The traditional methods on the OUDA problem
mainly focus on transforming each arriving target data to the source domain, 
and they do not sufficiently consider
the temporal coherency and accumulative statistics 
among the arriving target data.
We propose a multi-step framework for the OUDA problem, 
which institutes a novel method 
to compute the mean-target subspace 
inspired by the geometrical interpretation on the Euclidean space.
This mean-target subspace contains accumulative temporal information 
among the arrived target data.
Moreover, the transformation matrix
computed from the mean-target subspace
is applied to the next target data
as a preprocessing step,
aligning the target data closer to the source domain.
Experiments on four datasets 
demonstrated the contribution of each step in our proposed multi-step OUDA framework 
and its performance over previous approaches.
\end{abstract}

\footnotetext[1]{This work was supported in part by the National
Science Foundation under Grant IIS-1813935.
Any opinion, findings,
and conclusions or recommendations expressed in this material are
those of the authors and do not necessarily reflect the views of
the National Science Foundation.}
\footnotetext[2]{We also gratefully acknowledge the support of NVIDIA Corporation 
with donation of a Titan XP GPU used for this research.}



\section{Introduction}

Domain Adaptation (DA)~\cite{patel2015visual} aims to reduce the discrepancy 
between different distributions of the source and the target domains.
In particular, the Unsupervised Domain Adaptation (UDA) problem 
focuses on the study that the target data are completely unlabelled, 
which is more plausible assumption 
for the recognition tasks in the real world.

There have been many studies on the UDA problem. 
For one branch of the studies,  
Gong et al.~\cite{gong2012geodesic} and Fernando et al.~\cite{fernando2013unsupervised} 
assumed that the source and the target domains share 
the common low-dimensional subspace. 
For another branch of studies on the UDA problem, 
Long et al.~\cite{long2015learning} and Sun et al.~\cite{sun2016return}
directly minimized the discrepancy 
between the source and the target domains.
Furthermore, Zhang et al.~\cite{zhang2017joint} and Wang et al.~\cite{wang2018visual} 
combined the techniques in both branches.
Vascon et al.~\cite{vascon2019unsupervised} and Wulfmeier et al.~\cite{wulfmeier2018incremental} 
suggested a new technique for the UDA problem using 
Nash equilibrium~\cite{nash1950equilibrium} and Generative Adversarial Networks (GAN)~\cite{goodfellow2014generative}, respectively.

\begin{figure*}[!t]
  \centering
  \includegraphics[width=0.8\linewidth]{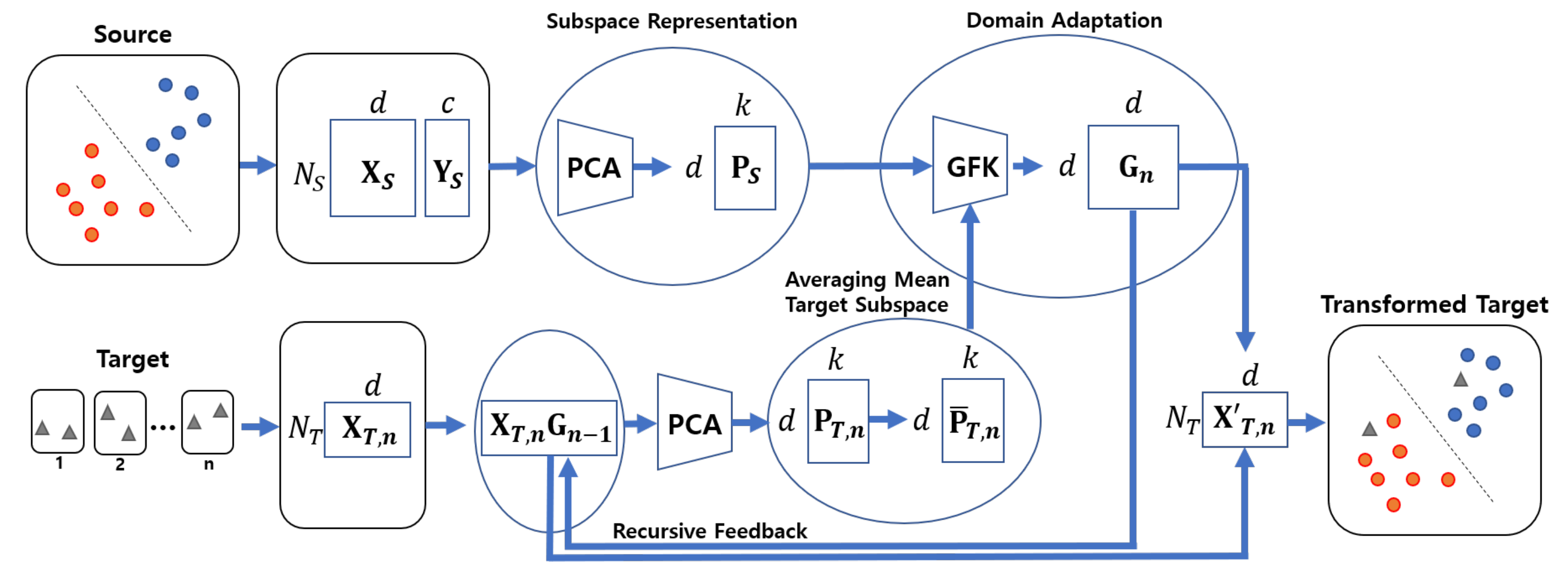}
  \caption{
The proposed OUDA method consists of four steps: 
  1) Subspace representation, 
  2) Averaging mean-target subspaces,
  3) Domain adaptation, and 
  4) Recursive feedback.}
  \label{fig:schematic}
\vspace*{-0.1in}
\end{figure*}

We notice that only a few work
has been conducted on the \textit{Online} Unsupervised Domain Adaptation (OUDA) problem, 
which assumes that the target data are arriving sequentially as a small batch.
Mancini et al.~\cite{mancini2018kitting} adopted 
a batch normalization technique~\cite{ioffe2015batch} 
for online domain adaptation, 
which was restricted to the kitting task only.
Wulfmeier et al.~\cite{wulfmeier2018incremental} expanded 
his previous work on GANs to the online case.
Bitarafan et al. proposed 
Incremental Evolving Domain Adaptation (IEDA)~\cite{bitarafan2016incremental} algorithm, 
which computes the target data transformation using Geodesic Flow Kernel (GFK)~\cite{gong2012geodesic} 
followed by updating the source subspace using 
Incremental Partial Least Square (IPLS)~\cite{zeng2014incremental}. 
This approach is vulnerable  
when the target data are predicted incorrectly 
because the ill-labelled target data would be merged with the source-domain data, 
leading to worse prediction of future target data.
Hoffman et al.~\cite{hoffman2014continuous} proposed an OUDA method 
using Continuous Manifold-based Adaptation (CMA), 
which formulated the OUDA problem 
as a non-convex optimization problem.
However, this method 
merely considered the coherency 
among the adjacent target-data batches.

To overcome the drawbacks of the above methods~--~
contamination of the source domain and lack of temporal coherency, 
we propose a multi-step framework for the OUDA problem, 
which institutes a novel method 
of computing the mean-target subspace
inspired by the geometrical interpretation 
in the Euclidean space.
Previous subspace-based methods on the OUDA problem 
merely compute the transformation matrix 
between the source subspace 
and each target subspace.
Our method instead computes the transformation matrix 
between the source subspace 
and the mean-target subspace, 
which is incrementally obtained 
on the Grassmann manifold.
Since a subspace is represented as a single point 
on the Grassmann manifold, 
the mean-target subspace is regarded 
as the mean point of multiple points 
that represent target subspaces 
of target-data batches.
Although Karcher mean~\cite{chikuse2012statistics} 
is a well-known method for computing the 
mean point on the Grassmann manifold, 
it is not suitable for the OUDA problem 
since the Karcher mean is computed 
with an iterative process.
Instead of the Karcher mean, 
we propose to compute the mean-target subspace 
by a geometrical process, 
which resembles the process of 
incremental computation for the mean point 
of a given multiple points on the Euclidean space.
The transformation matrix 
computed with our proposed method
is robust to the abrupt change 
of arriving target batches, 
leading to a stable domain transfer.
We also feed the transformation matrix 
back to the next target batch, 
which moves it closer to the source domain.
This preprocessing step of the next target batch 
leads to a more precise computation
of the mean-target subspace.
Experiments on four datasets demonstrated that 
our proposed method outperforms 
the traditional methods in terms of 
performance and computation speed.

\section{Proposed Approach}
\subsection{Problem Description}
We assume that the data in the source domain 
$\mathbf{X}_{\mathcal{S}}\in \mathbb{R}^{N_{\mathcal{S}} \times d}$ 
are static and labelled as 
$\mathbf{Y}_{\mathcal{S}}\in \mathbb{R}^{N_{\mathcal{S}} \times c}$, 
where $N_{\mathcal{S}}$, $d$ and $c$ 
indicate the numbers of source data, dimension of the data 
and the number of the class categories, respectively.
Data in the target domain 
are unlabelled 
and arriving as one batch in each timestep as 
$\mathbf{X}_{\mathcal{T}}=\{\mathbf{X}_{\mathcal{T},1}, \mathbf{X}_{\mathcal{T},2}, \ldots, \mathbf{X}_{\mathcal{T},B}\}$, 
which are assumed to be sequential and temporally correlated.
We use the term \textit{mini-batch} for the $n^{th}$ target-data batch
$\mathbf{X}_{\mathcal{T},n}\in \mathbb{R}^{N_{\mathcal{T}} \times d}$
and $B$ indicates the number of mini-batches 
and $N_{\mathcal{T}}$ indicates the number of data in each mini-batch.
$N_{\mathcal{T}}$ is assumed to be constant for $n=1,2,\ldots,B$ 
and very small compared to $N_\mathcal{S}$.
In our notation, the subscripts $\mathcal{S}$ and $\mathcal{T}$ 
indicate the source and the target domains, respectively.
Furthermore, subscript $(\mathcal{T},n)$ represents 
the $n^{th}$ mini-batch in the target domain.

Our goal is to align the target-data batch $\mathbf{X}_{\mathcal{T},n}$ 
to the source domain at 
$n=1,2,\ldots,B$ 
in an online manner 
so that the transformed target data 
$\mathbf{X'}_{\mathcal{T},n}$
can be recognized correctly as $\mathbf{\hat{Y}}_{\mathcal{T},n}$ 
with the classifier pre-trained in the source domain. 
Using the notation of~\cite{gong2012geodesic}, 
we denote the subspace with its basis $\mathbf{P}_{\mathcal{S}}$ 
and $\mathbf{P}_{\mathcal{T},n}\in \mathbb{R}^{d \times k}$, 
where $d$ is the dimension of the original data 
and $k$ is the dimension of the subspace.
For instance, 
$\mathbf{P}_{\mathcal{T}}=\{\mathbf{P}_{\mathcal{T},1}, \mathbf{P}_{\mathcal{T},2}, \ldots , \mathbf{P}_{\mathcal{T},B}\}$ 
is the set of target subspaces composed of entire mini-batches, 
whereas $\mathbf{P}_{\mathcal{T},n}$ is the target subspace for the $n^{th}$ mini-batch.
For example, for Principal Component Analysis (PCA)~\cite{wold1987principal}, 
this subspace represents the projection matrix 
from the original space to the subspace.

\subsection{Proposed OUDA Method}
As shown in Fig.~\ref{fig:schematic}, 
our proposed OUDA framework  
consists of four steps 
for the incoming $n^{th}$ mini-batch: 
1) Subspace representation, 
2) Averaging mean-target subspace, 
3) Domain adaptation, 
and 4) Recursive feedback.
Step one computes the low-dimensional subspace,
$\mathbf{P}_{\mathcal{T},n}$, of the target domain using PCA.
Step two computes the mean of the target subspaces 
$\mathbf{\overline{P}}_{\mathcal{T},n}$ 
embedded in the Grassmann manifold 
using our novel technique, \textit{Incremental Computation of Mean Subspace (ICMS)}.
Step three is the domain adaptation 
and it computes the transformation matrix $\mathbf{G}_{n}$ 
from the target domain to the source domain 
based on the approach of Bitarafan et al. \cite{bitarafan2016incremental} 
which adopts the GFK method \cite{gong2012geodesic}, 
a manifold alignment technique.
Step four provides recursive feedback 
by feeding $\mathbf{G}_{n}$ back to the next mini-batch 
$\mathbf{X}_{\mathcal{T},n+1}$.
Each step is described next in detail.

\subsubsection{Subspace Representation}
The ultimate goal of our proposed OUDA method is to find 
the transformation matrix 
$\mathbf{G}=\{\mathbf{G}_{1}, \mathbf{G}_{2}, \ldots , \mathbf{G}_{B}\}$ 
that transforms the set of target mini-batches 
$\mathbf{X}_{\mathcal{T}}=\{\mathbf{X}_{\mathcal{T},1}, \mathbf{X}_{\mathcal{T},2}, \ldots , \mathbf{X}_{\mathcal{T},B}\}$
to 
$\mathbf{X'}_{\mathcal{T}}=\{\mathbf{X'}_{\mathcal{T},1}, \mathbf{X'}_{\mathcal{T},2}, \ldots , \mathbf{X'}_{\mathcal{T},B}\}$ 
so that these transformed target data
are well aligned to the source domain, 
where $\mathbf{G}_{n}\in\mathbb{R}^{d \times d}$ indicates 
the transformation matrix from $\mathbf{X}_{\mathcal{T},n}$ 
to $\mathbf{X'}_{\mathcal{T},n}$.
However, we prefer not to use the methods 
that compute $\mathbf{G}_{n}$ directly on the original data space 
with high dimension $d$.
For example, raw input image features have dimension $d=4096$, 
and the technique, which directly computes the transformation matrix 
by Correlation Alignment (CORAL)~\cite{sun2016return}, 
requires to compute $4096\times 4096$ matrix.
Since our technique is desired to be conducted in online manner, 
we embed the source $\mathbf{X}_{\mathcal{S}}$ 
and the target data $\mathbf{X}_{\mathcal{T}}=\{\mathbf{X}_{\mathcal{T},1}, \mathbf{X}_{\mathcal{T},2}, \ldots , \mathbf{X}_{\mathcal{T},B}\}$ 
to low-dimensional spaces as $\mathbf{P}_{\mathcal{S}}$ and $\mathbf{P}_{\mathcal{T}}=\{\mathbf{P}_{\mathcal{T},1}, \mathbf{P}_{\mathcal{T},2}, \ldots , \mathbf{P}_{\mathcal{T},B}\}$, respectively, 
which preserve the meaningful information of the original data space.
We adopt PCA to obtain $\mathbf{P}_{\mathcal{S}}$ and $\mathbf{P}_{\mathcal{T}}$ 
since PCA algorithm is simple and fast for online DA, 
and it is available for both labellel and unlabelled data 
unlike other dimension-reduction techniques 
such as Linear Discriminant Analysis (LDA) \cite{balakrishnama1998linear}.

\subsubsection{Averaging Mean-target Subspace}
Throughout this paper, 
we utilize Grassmann manifold $G(k, d)$~\cite{edelman1998geometry}, 
a space that parameterizes all $k$ dimensional linear subspaces 
of $d$ dimensional vector space. 
Since a subspace is represented as a single point on the Grassmann manifold, 
$\mathbf{P}_{\mathcal{S}}$ and $\mathbf{P}_{\mathcal{T},1}, \mathbf{P}_{\mathcal{T},2}, \ldots , \mathbf{P}_{\mathcal{T},B}$ 
are represented as $(B+1)$ points on $G(k, d)$.

For solving the offline UDA problem (i.e., $B=1$), 
Gong et al.~\cite{gong2012geodesic} utilized the geodesic flow 
from $\mathbf{P}_{\mathcal{S}}$ to $\mathbf{P}_{\mathcal{T}}$ on $G(k, d)$.
Previous methods for the OUDA problem directly compute 
the transformation matrix based on the source and the target subspaces 
of each mini-batch.
We propose a novel technique, 
called \textit{Incremental Computation of Mean Subspace (ICMS)}, 
which computes the mean subspace in the target domain 
inspired by the geometrical interpretation on the Euclidean space.
Then we compute the geodesic flow from 
$\mathbf{P}_{\mathcal{S}}$ to $\mathbf{\overline{P}}_{\mathcal{T}, n}$.
Formally, when the $n^{th}$ mini-batch $\mathbf{X}_{\mathcal{T}, n}$ 
arrives and is represented as the subspace $\mathbf{P}_{\mathcal{T}, n}$, 
we incrementally compute the mean-target subspace 
$\mathbf{\overline{P}}_{\mathcal{T},n}$ 
using $\mathbf{P}_{\mathcal{T}, n}$ 
and $\mathbf{\overline{P}}_{\mathcal{T},n-1}$, 
where $\mathbf{\overline{P}}_{\mathcal{T},n-1}$ is the mean subspace 
of ($n-1$) target subspaces 
$\mathbf{P}_{\mathcal{T},1}, \mathbf{P}_{\mathcal{T},2}, \ldots , \mathbf{P}_{\mathcal{T},n-1}$.

\begin{figure}[!t]
  \centering
  \begin{subfigure}[t]{0.3767\linewidth}
    \includegraphics[width=\linewidth]{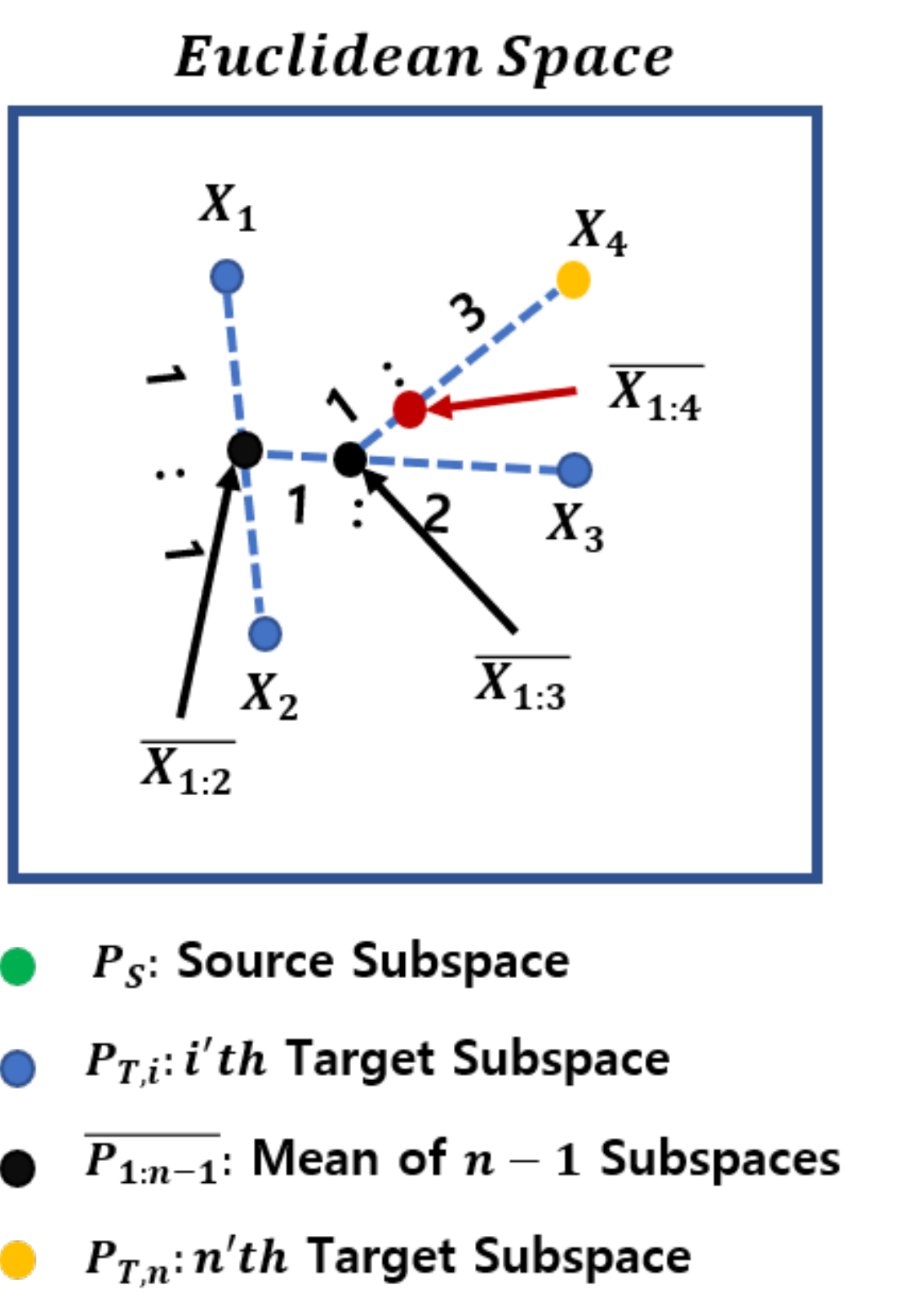}
    \caption{Euclidean Space}
    \label{fig:Euclidean}
  \end{subfigure}
  \begin{subfigure}[t]{0.5232\linewidth}
    \includegraphics[width=\linewidth]{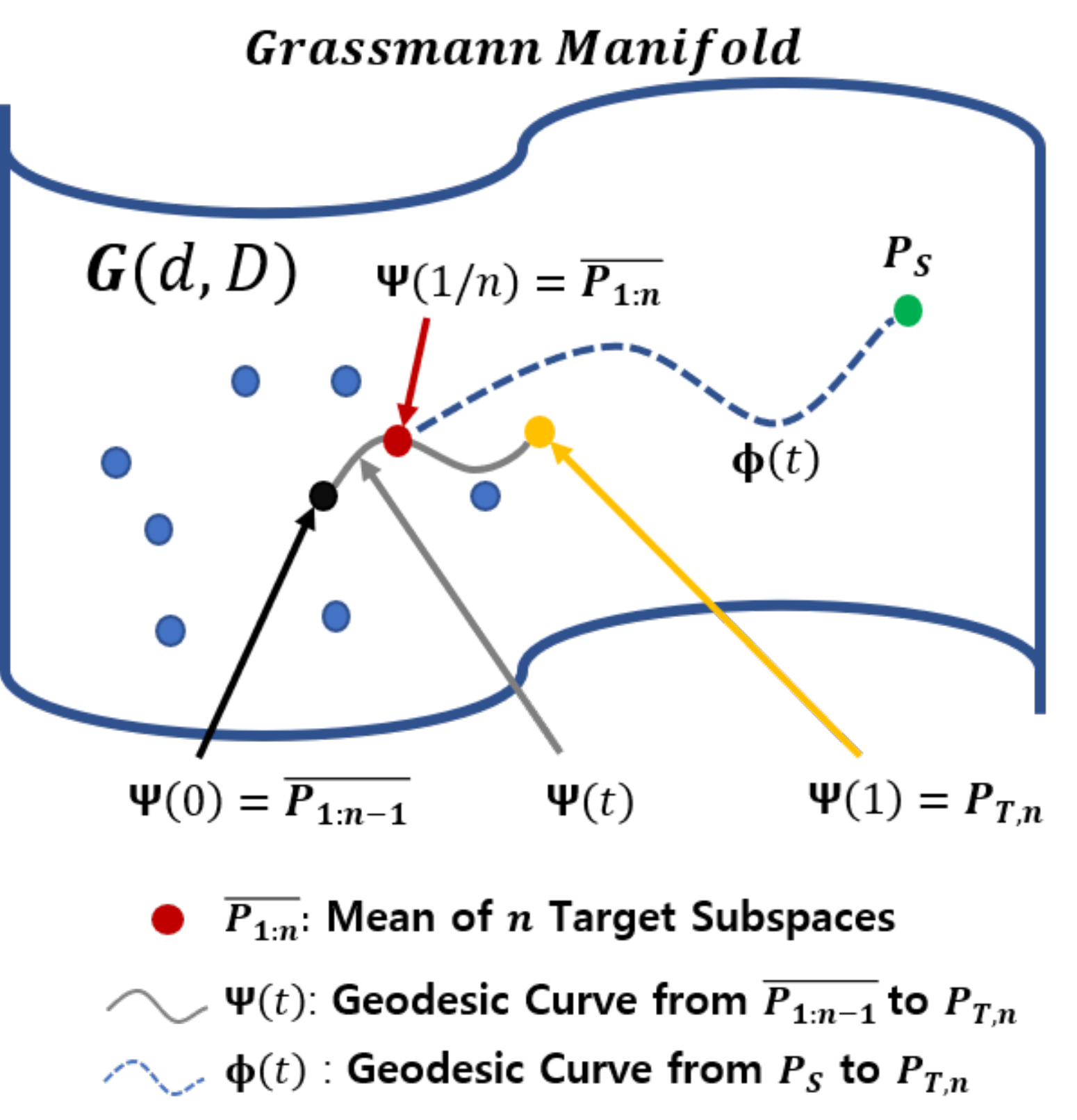}
    \caption{Grassmann Manifold}
    \label{fig:Grassmann}
  \end{subfigure}
  \caption{(a) Incremental computation of mean-target subspace  
  inspired by the geometrical interpretation 
  from Euclidean space.
 (b) Incremental computation of the mean of target subspaces 
  on the Grassmann manifold.}
  \label{fig:subspace_avg2}
\end{figure}

As shown in Fig.~\ref{fig:subspace_avg2}(a), 
the mean point 
$\mathbf{\overline{X}}_{n}$ can be computed in an incremental way 
when $n$ points $\mathbf{X}_{1}, \mathbf{X}_{2}, \ldots , \mathbf{X}_{n}$ 
are on the Euclidean space.
If the mean point $\mathbf{\overline{X}}_{n-1}$ of $n-1$ points $\mathbf{X}_{1}, \mathbf{X}_{2}, \ldots , \mathbf{X}_{n-1}$ and the $n^{th}$ point $\mathbf{X}_{n}$ are given, the updated mean point $\mathbf{\overline{X}}_{n}$ is computed as $\mathbf{\overline{X}}_{n}=\{(n-1)\mathbf{\overline{X}}_{n-1}+\mathbf{X}_{n}\}/n$.
From a geometrical perspective, 
$\mathbf{\overline{X}}_{n}$ is the internal point where the distances from 
$\mathbf{\overline{X}}_{n}$ to $\mathbf{\overline{X}}_{n-1}$ 
and to $\mathbf{X}_{n}$ have the ratio of $1:(n-1)$:
\begin{equation}
\label{eq:Euclidean}
    |\mathbf{\overline{X}}_{n-1}\mathbf{\overline{X}}_{n}|
    =\frac{|\mathbf{\overline{X}}_{n-1}\mathbf{X}_{n}|}{n}.    
\end{equation}

We adopt this ratio concept  
to the Grassmann manifold 
from a geometrical perspective. 
As shown in Fig.~\ref{fig:subspace_avg2}(b), 
we update the mean-target subspace 
$\mathbf{\overline{P}}_{\mathcal{T},n}$ 
of $n$ target subspaces when the previous mean 
subspace $\mathbf{\overline{P}}_{\mathcal{T},n-1}$ 
of ($n-1$) target subspaces 
and $n^{th}$ subspace $\mathbf{P}_{\mathcal{T},n}$ are given.
Using the geodesic parameterization~\cite{gallivan2003efficient} 
with a single parameter $t$, 
the geodesic flow from 
$\mathbf{\overline{P}}_{\mathcal{T},n-1}$ 
to $\mathbf{P}_{\mathcal{T},n}$ 
is parameterized as 
$\mathbf{\Psi}_{n}:t\in[0,1]\xrightarrow{}\mathbf{\Psi}_{n}(t)\in G(k,d)$:
\begin{equation}
\label{eq:ICMS}
    \mathbf{\Psi}_{n}(t)=\mathbf{\overline{P}}_{\mathcal{T},n-1}\mathbf{U}_{1,n}\mathbf{\Gamma}_{n}(t) - \mathbf{\overline{R}}_{\mathcal{T},n-1}\mathbf{U}_{2,n}\mathbf{\Sigma}_{n}(t)
\end{equation}
under the constraints 
$\mathbf{\Psi}_{n}(0)=\mathbf{\overline{P}}_{\mathcal{T},n-1}$ and $\mathbf{\Psi}_{n}(1)=\mathbf{P}_{\mathcal{T},n}$.
It is valid to apply this ratio concept on the Euclidean space 
to the geodesic flow on the Grassmann manifold 
since $t$ is parameterized proportionally to the arc length 
of $\mathbf{\Psi}_{n}(t)$~\cite{robbin2011introduction}.
$\mathbf{\overline{R}}_{\mathcal{T},n-1}\in\mathbb{R}^{d\times(d-k)}$ 
denotes the orthogonal complement to $\mathbf{\overline{P}}_{\mathcal{T},n-1}$; 
that is, $\mathbf{\overline{R}}_{\mathcal{T},n-1}^{T}\mathbf{\overline{P}}_{\mathcal{T},n-1}=\mathbf{O}$.
Two orthonormal matrices 
$\mathbf{U}_{1,n}\in \mathbb{R}^{k \times k}$ 
and $\mathbf{U}_{2,n}\in \mathbb{R}^{(d-k) \times (d-k)}$ 
are given by the following pair of singular-value decompositions (SVDs),

\begin{align}
\label{eq:SVD_mean}
    \mathbf{\overline{P}}_{\mathcal{T},n-1}^{T}\mathbf{P}_{\mathcal{T},n}=\mathbf{U}_{1,n}\mathbf{\Gamma}_{n}\mathbf{V}_{n}^{T} \;\:\, \\ 
    \mathbf{\overline{R}}_{\mathcal{T},n-1}^{T}\mathbf{P}_{\mathcal{T},n}=-\mathbf{U}_{2,n}\mathbf{\Sigma}_{n}\mathbf{V}_{n}^{T}
\end{align}
where $\mathbf{\Gamma}_{n} \in \mathbb{R}^{k \times k}$ and $\mathbf{\Sigma}_{n}=[\mathbf{\Sigma}_{1,n}^{T} \text{     } \mathbf{O}^{T}]^{T} \in \mathbb{R}^{(d-k) \times k}$ 
are diagonal and block diagonal matrices, respectively, 
and $\mathbf{\Sigma}_{1,n} \in \mathbb{R}^{k \times k}$ 
and $\mathbf{O} \in \mathbb{R}^{(d-2k) \times k}$. 
Since the dimension of $\mathbf{O}$ should be positive, 
$(d-2k)$ should be greater than 0. 
We assume that the dimension of the subspace $k$ 
is much smaller 
than the dimension of the original space $d$ 
so that $k < d/2$. 
The diagonal elements of 
$\mathbf{\Gamma}_{n}$ and $\mathbf{\Sigma}_{1,n}$ 
are $\cos{\theta_{i,n}}$ and $\sin{\theta_{i,n}}$ for $i=1,2,\ldots ,k$. 
These $\theta_{i,n}$'s are the principal angles~\cite{knyazev2012principal} 
between $\mathbf{\overline{P}}_{\mathcal{T},n-1}$ 
and $\mathbf{P}_{\mathcal{T},n}$.
$\mathbf{\Gamma}_{n}(t)$ and $\mathbf{\Sigma}_{n}(t)=[\mathbf{\Sigma}_{1,n}(t)^{T} \text{     } \mathbf{O}^{T}]^{T}$ 
are diagonal and block diagonal matrices 
whose elements are $\cos{(t\theta_{i,n})}$ 
and $\sin{(t\theta_{i,n})}$, respectively.

Finally, we adopt the ratio concept from Eq. \eqref{eq:Euclidean} 
to $\mathbf{\Psi}_{n}(t)$ and obtain $\mathbf{\overline{P}}_{\mathcal{T},n}=\mathbf{\Psi}_{n}(\frac{1}{n})$.
Hence, we can incrementally compute the mean-target subspace as follow:
\begin{equation}
\label{eq:mean_target_subspace}
    \mathbf{\overline{P}}_{\mathcal{T},n}=\mathbf{\overline{P}}_{\mathcal{T},n-1}\mathbf{U}_{1,n}\mathbf{\Gamma}_{n}(\frac{1}{n}) - \mathbf{\overline{R}}_{\mathcal{T},n-1}\mathbf{U}_{2,n}\mathbf{\Sigma}_{n}(\frac{1}{n}).    
\end{equation}
Note that $n$ refers to the $n^{th}$ mini-batch in the target domain. 
Since $0 \leq \frac{1}{n} \leq 1$,  
$\mathbf{\Gamma}_{n}(\frac{1}{n})$ and $\mathbf{\Sigma}_{n}(\frac{1}{n})$ 
are well defined.

\subsubsection{Domain Adaptation}

After computing the mean-target subspace 
$\mathbf{\overline{P}}_{\mathcal{T},n}$, 
we parameterize another geodesic flow 
from $\mathbf{P}_{\mathcal{S}}$ 
to $\mathbf{\overline{P}}_{\mathcal{T},n}$ 
as $\mathbf{\Phi}_{n}:t\in[0,1]\xrightarrow{}\mathbf{\Phi}_{n}(t)\in G(k,d)$:

\begin{equation}
\label{eq:GFK}
    \mathbf{\Phi}_{n}(t)=\mathbf{P}_{\mathcal{S}}\mathbf{U}_{3,n}\mathbf{\Lambda}_{n}(t) - \mathbf{R}_{\mathcal{S}}\mathbf{U}_{4,n}\mathbf{\Omega}_{n}(t)
\end{equation}
under the constraints $\mathbf{\Phi}_{n}(0)=\mathbf{P}_{\mathcal{S}}$ 
and $\mathbf{\Phi}_{n}(1)=\mathbf{\overline{P}}_{\mathcal{T},n}$. $\mathbf{R}_{\mathcal{S}}\in\mathbb{R}^{d\times(d-k)}$ denotes the orthogonal complement 
to $\mathbf{P}_{\mathcal{S}}$; that is,  $\mathbf{R}_{\mathcal{S}}^{T}\mathbf{P}_{\mathcal{S}}=\mathbf{0}$.
Two orthonormal matrices $\mathbf{U}_{3,n}\in \mathbb{R}^{k \times k}$ and $\mathbf{U}_{4,n}\in \mathbb{R}^{(d-k) \times (d-k)}$ 
are given by the following pair of SVDs,
\begin{align}
\label{eq:SVD_GFK}
    \mathbf{P}_{\mathcal{S}}^{T}\mathbf{\overline{P}}_{\mathcal{T},n}=\mathbf{U}_{3,n}\mathbf{\Lambda}_{n}\mathbf{W}_{n}^{T} \;\:\, \\ 
    \mathbf{R}_{\mathcal{S}}^{T}\mathbf{\overline{P}}_{\mathcal{T},n}=-\mathbf{U}_{4,n}\mathbf{\Omega}_{n}\mathbf{W}_{n}^{T} .
\end{align}
Based on the GFK, 
the transformation matrix $\mathbf{G}_{n}$ 
from the target domain 
to the source domain 
is found by projecting and integrating 
over the infinite set 
of all intermediate subspaces between them:
\begin{equation}
    \int_{0}^{1}(\mathbf{\Phi}_{n}(\alpha)^{T}\mathbf{x}_{i})^{T}(\mathbf{\Phi}_{n}(\alpha)^{T}\mathbf{x}_{j})d\alpha=\mathbf{x}_{i}^{T}\mathbf{G}_{n}\mathbf{x}_{j} .
\end{equation}
From the above equation, we can derive the closed form of $\mathbf{G}_{n}$ as:
\begin{equation}
    \mathbf{G}_{n}=\int_{0}^{1}\mathbf{\Phi}_{n}(\alpha)\mathbf{\Phi}_{n}(\alpha)^{T}d\alpha .
\end{equation}
We adopt this $\mathbf{G}_{n}$ 
as the transformation matrix 
to the preprocessed target data as $\mathbf{X'}_{\mathcal{T},n}=\mathbf{X}^{pre}_{\mathcal{T},n}\mathbf{G}_{n}$, 
which better aligns the target data to the source domain.
$\mathbf{X}^{pre}_{\mathcal{T},n}$ is the target data 
fed back from the previous mini-batch, 
which is described in the next section.

\subsubsection{Recursive Feedback}
Previous work on the OUDA problem 
does not evidently consider the temporal dependency 
between the subspace of adjacent target mini-batches.
Unlike traditional methods, our proposed OUDA method feeds 
$\mathbf{G}_{n}$ back to the next target mini-batch 
as $\mathbf{X}^{pre}_{\mathcal{T},n+1}
=\mathbf{X}_{\mathcal{T},n+1}\mathbf{G}_{n}$ 
at the next timestep ($n+1$), 
which imposes the temporal dependency 
between $\mathbf{X}_{\mathcal{T},n}$ and $\mathbf{X}_{\mathcal{T},n+1}$ 
by moving $\mathbf{P}_{\mathcal{T},n+1}$ 
closer to $\mathbf{P}_{\mathcal{T},n}$ 
on the Grassmann manifold. 
PCA is conducted from this $\mathbf{X}^{pre}_{\mathcal{T},n+1}$ 
to represent the $(n+1)^{th}$ target subspace $\mathbf{P}_{\mathcal{T},n+1}$.

\section{Experimental Results}
\subsection{Datasets}
\begin{figure}[!t]
  \centering
    \includegraphics[width=\linewidth]{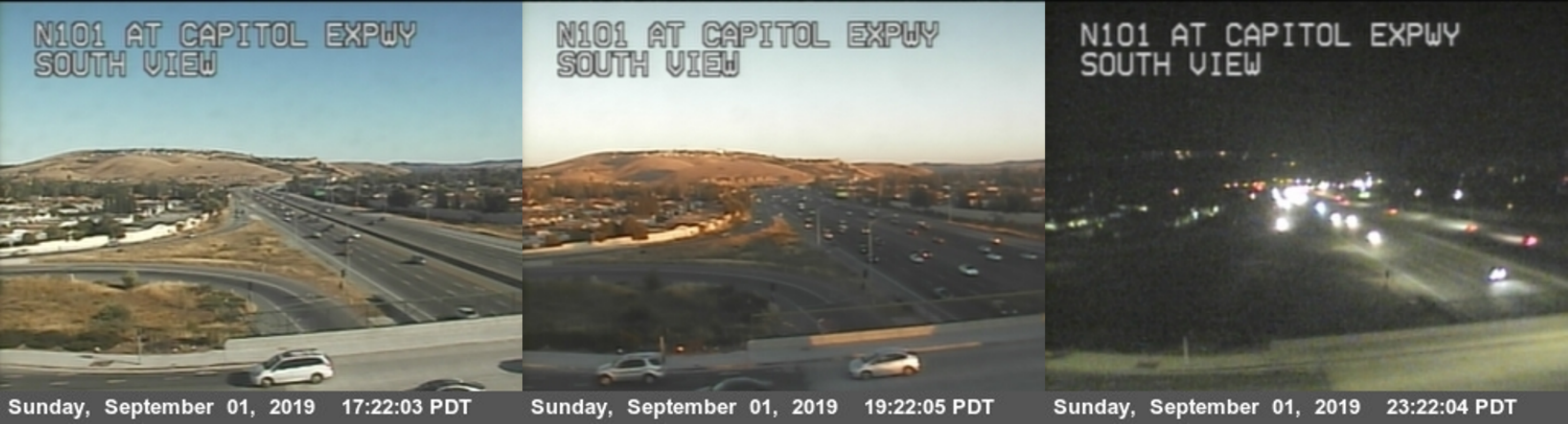}
  \caption{Image samples of Traffic dataset 
  captured from the morning (left) 
  to afternoon (middle) and night (right).}
  \label{fig:dataset}
\end{figure}

To evaluate our proposed OUDA method in data classification, 
we performed experiments on four datasets~\cite{bitarafan2016incremental}-- 
the Traffic dataset, 
the Car dataset,
the Waveform21 dataset, and
the Waveform40 dataset. 
These datasets provided a large variety of time-variant images 
and signals to test upon. 
The Traffic dataset includes images captured from 
a fixed traffic camera observing a road over a 2-week period. 
It consists of 5412 instances of $d=512$ dimensional features 
with two classes as either heavy traffic or light traffic.
Figure~\ref{fig:dataset}
depicts the image samples of the Traffic dataset.
The Car dataset contains images of automobiles manufactured 
between 1950 and 1999 acquired from online database. 
It includes 1770 instances of $d=4096$ dimensional features 
with two classes as sedans or trucks.  
The Waveform21 dataset is composed of 5000 wave instances 
of $d=21$ dimensional features with three classes.   
The Waveform40 dataset is the second version of the Waveform21 
with additional features. 
This dataset consists of $d=40$ dimensional features.

\subsection{Comparison with Previous Methods}
We used the Evolving Domain Adaptation (EDA)~\cite{bitarafan2016incremental} method 
as the reference model for comparing the classification accuracy 
with our proposed OUDA method and its variants.
The metric for classification accuracy 
is based on~\cite{bitarafan2016incremental} as 
$A(n)=\{\sum_{\tau =1}^{n}a(\tau)\}/n$, 
where $A(n)$ is the accuracy of the $n$ arrived data 
and $a(\tau)$ is the accuracy for $\tau^{th}$ mini-batch.

\begin{figure}[!t]
  \centering
  \begin{subfigure}[b]{0.49\linewidth}
    \includegraphics[width=\linewidth]{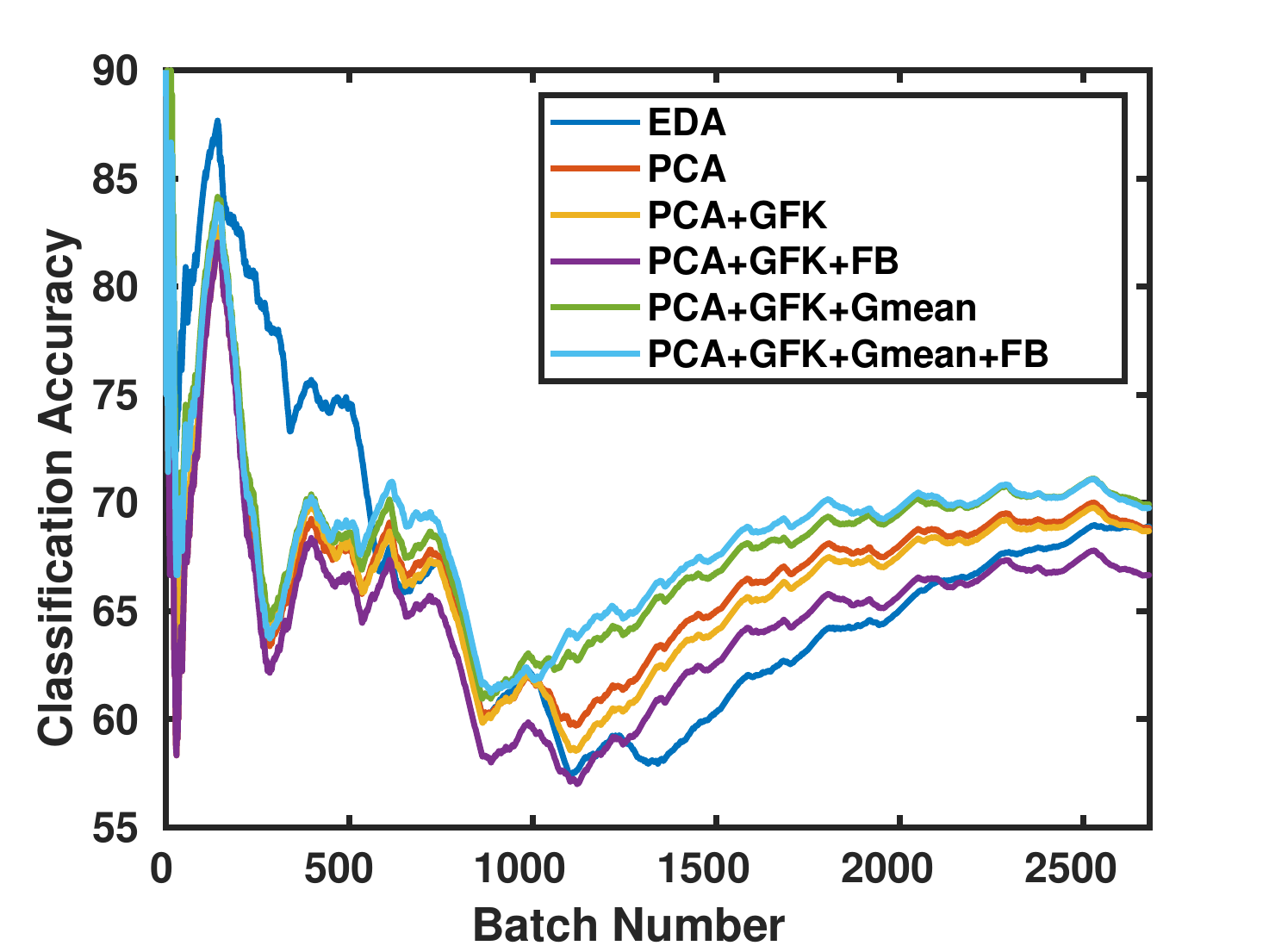}
    \caption{Traffic}
    \label{fig:plot_traffic}
  \end{subfigure}
  \begin{subfigure}[b]{0.49\linewidth}
    \includegraphics[width=\linewidth]{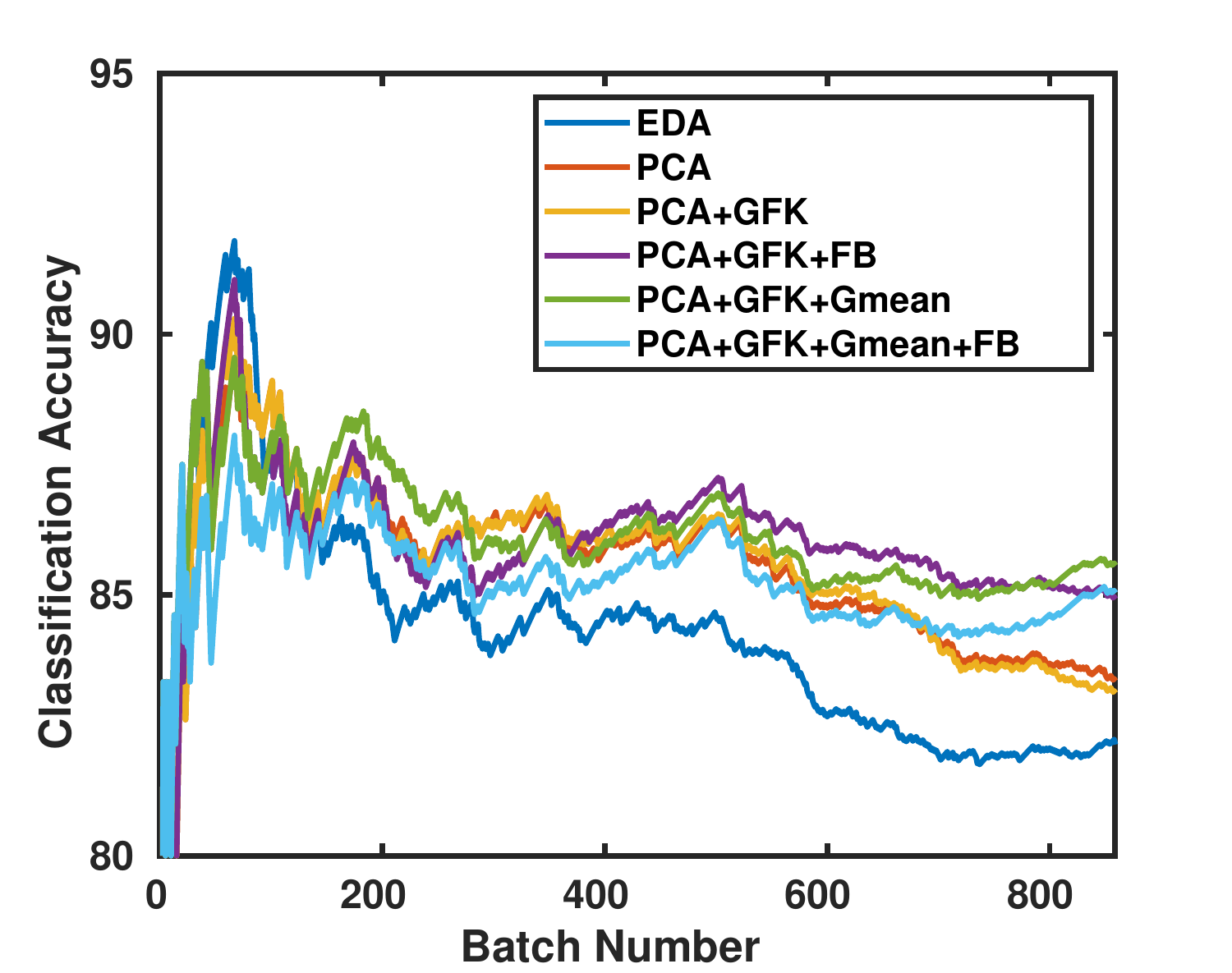}
    \caption{Car}
    \label{fig:plot_car}
  \end{subfigure}
  
  \begin{subfigure}[b]{0.49\linewidth}
    \includegraphics[width=\linewidth]{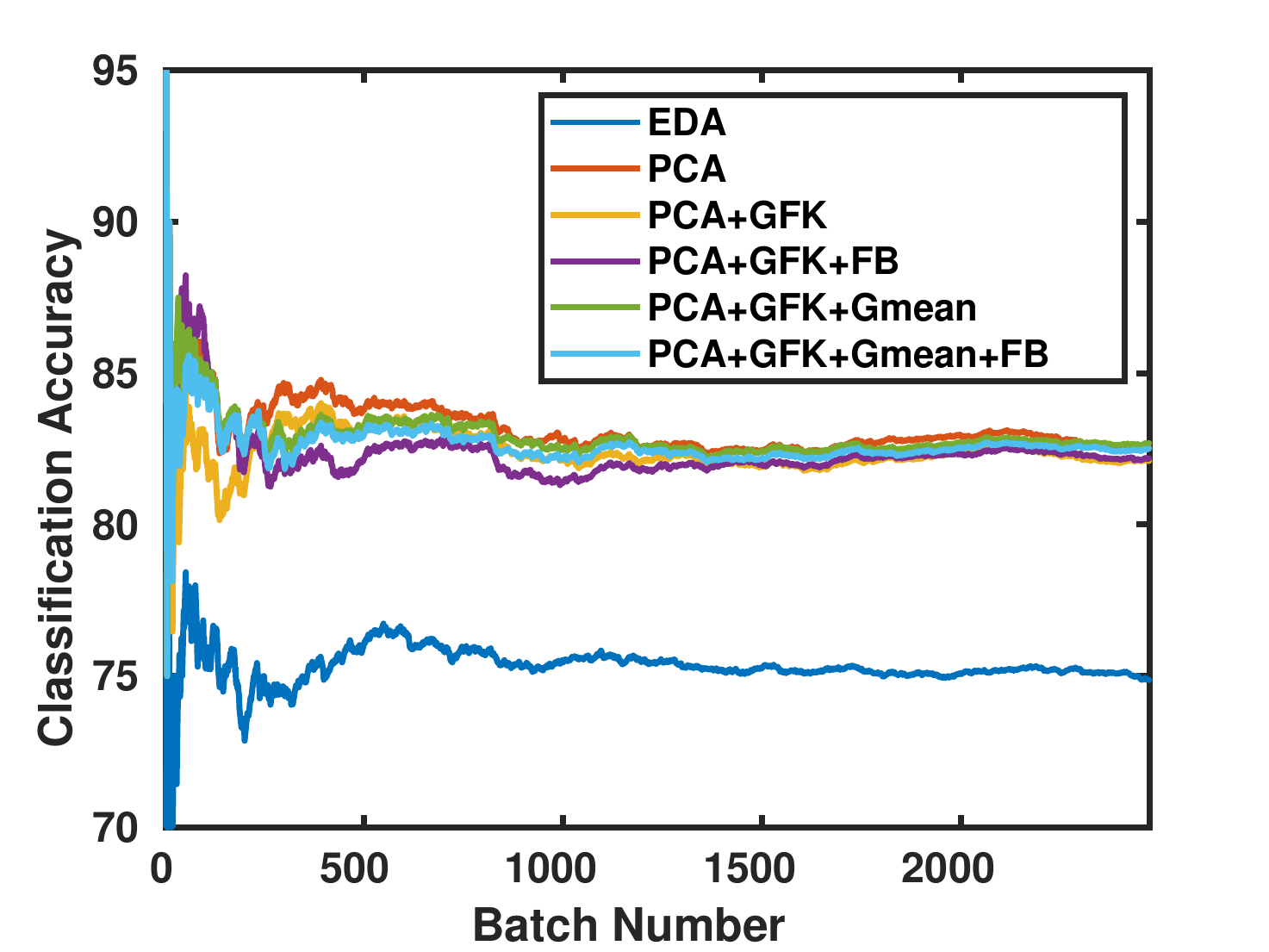}
    \caption{Waveform21}
    \label{fig:wave21}
  \end{subfigure}
  \begin{subfigure}[b]{0.49\linewidth}
    \includegraphics[width=\linewidth]{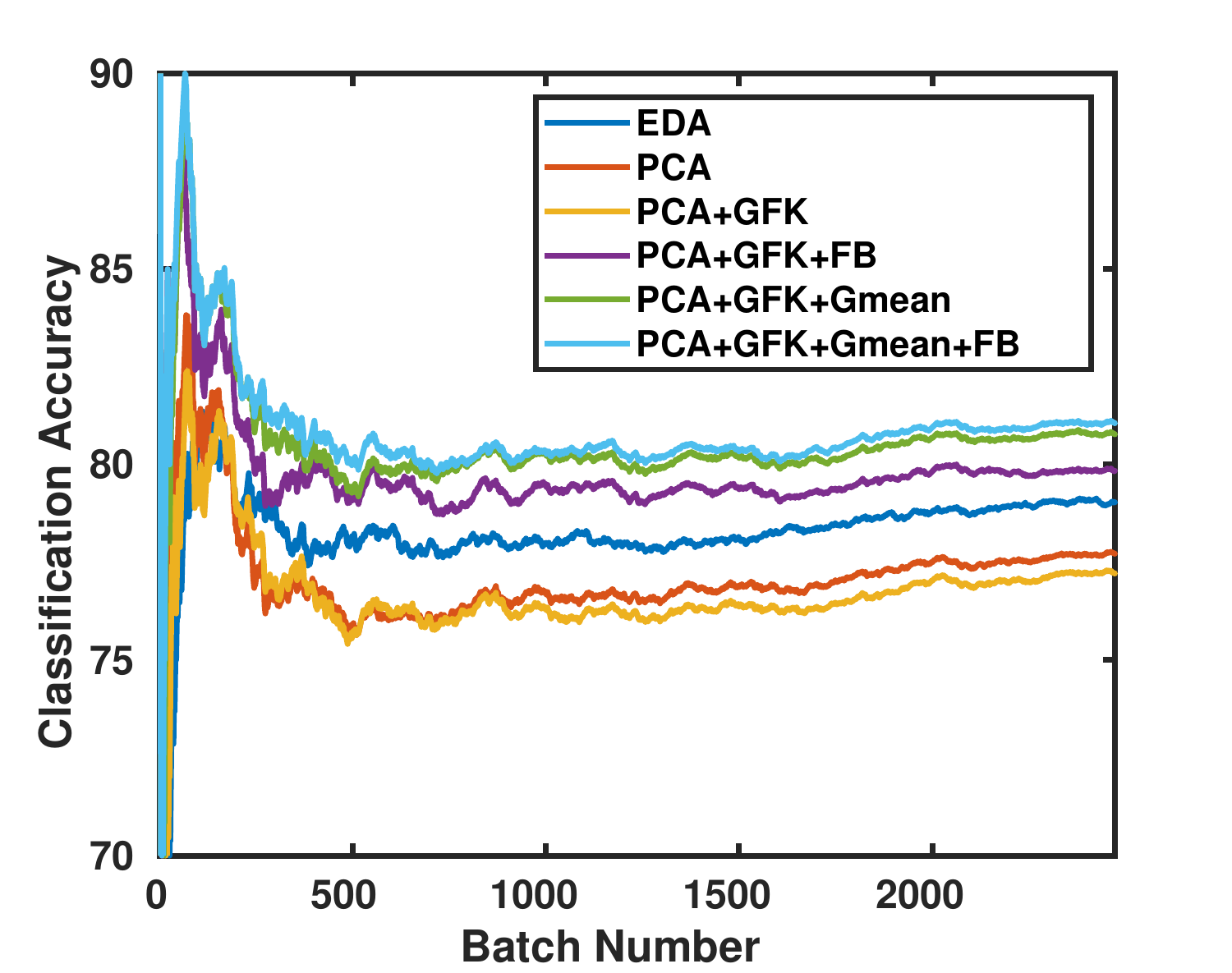}
    \caption{Waveform40}
    \label{fig:wave40}
  \end{subfigure}
  \caption{Accuracy of the previous method (EDA) 
  and variants of the proposed method.}
  \label{fig:plot}
\end{figure}

Figure~\ref{fig:plot} depicts the classification accuracy 
when the mini-batches are arriving.
It indicated that our proposed OUDA method and majority of its variants 
outperformed the EDA method.
For the Traffic dataset, a sudden drift occurred in the $1100^{th}$ 
mini-batch which resulted in an abrupt decrease of the accuracy 
but the performance recovered 
when the number of arriving mini-batch increased.
For the Car dataset, the average accuracy was slightly decreasing 
since the target data were evolving in long term (i.e., from 1950 to 1999), 
which resulted in more discrepancy between the source and the target domains.

\subsection{Ablation Study}
In order to understand which step of our proposed method
contributes to the improvement of the accuracy performance, 
we also measured the accuracy for the different variants of 
our proposed OUDA method and compared their performance.
We compared the accuracy by incrementally 
including each step to the process of OUDA.
Except for the EDA method, 
which adopted Incremental Semi-Supervised Learning (ISSL) technique 
for classifying the unlabelled target data, 
all other approaches 
adopted the basic K-Nearest-Neighbors \cite{keller1985fuzzy} 
or Support-Vector-Machine \cite{suykens1999least} classifiers 
for target-label prediction.

Table \ref{tab:shallow} shows that averaging the mean-target subspace (Gmean) 
and recursive feedback (FB) steps improved the performance the most.
Gmean and FB steps improved the performance at 4.27\% and 4.08\% respectively, compared to EDA.
These results indicated that computing the mean-target subspace
leads to stable computation of the transformation matrix $\mathbf{G}_{n}$.
Furthermore, feeding $\mathbf{G}_{n}$ 
back to the $(n+1)^{th}$ target mini-batch 
shifted it closer to the source domain.

\begin{table}[!t]
 \caption{Accuracy (\%) of Various Methods(Vanilla AE)}
  \centering
  \resizebox{\columnwidth}{!}{%
  \begin{tabular}{lccccc}
  \hline
  Method & Classifier & Traffic & Car & Waveform21 & Waveform40 \\
  \hline
  CMA+GFK & KNN & 63.22 & 82.50 & 72.48 & 66.85 \\
  & SVM & 68.87 & 82.73 & 69.15 & 68.77  \\
  CMA+SA & KNN & 41.33 & 56.45 & 33.19 & 33.09  \\
  & SVM & 41.33 & 56.45 & 33.84 & 33.05 \\
  EDA & ISSL & 69.00 & 82.59 & 74.65 & 79.66  \\
  \hline
  PCA & KNN & 63.05 & 82.50 & 71.07 & 66.08 \\
  & SVM & 68.85 & 83.31 & 82.55 & 77.74 \\
  PCA+GFK & KNN & 64.02 & 82.44 & 70.55 & 65.76 \\
  & SVM & 68.71 & 83.08 & 82.10 & 77.23 \\
  PCA+GFK+FB & KNN & 61.77 & 81.28 & 72.65 & 66.85 \\
  & SVM & 66.67 & 84.88 & 82.18 & 79.86 \\
  PCA+GFK+Gmean & KNN & 56.42 & 82.73 & 72.22 & 67.11\\
  & SVM & \textbf{69.94} & \textbf{85.52} & \textbf{82.69} & 80.79 \\
  PCA+GFK+Gmean+FB & KNN & 57.03 & 82.44 & 72.38 & 67.90\\
  & SVM & 69.77 & 85.00 & 82.51 & \textbf{81.07} \\

  \hline
  \end{tabular}
  }
  \label{tab:shallow}
\end{table}
\begin{table}[!t]
 \caption{Comparison of Computation Time (sec)}
  \centering
  \resizebox{\columnwidth}{!}{%
  \begin{tabular}{lcccc}
  \hline
  Method & Traffic & Car & Waveform21 & Waveform40\\
  \hline
  EDA & 105.7 & 2545 & 22.32 & 23.42\\
  Proposed method & 57.45 & 5503 & 3.188 & 4.410 \\
  \hline
  \end{tabular}
  }
  \label{tab:time}
\end{table}

\subsection{Computation Time}
We evaluated the computation time of our proposed OUDA method 
as compared to the previous methods in the same datasets above.
As shown in Table \ref{tab:time}, 
our proposed OUDA method was significantly 
faster (i.e, 1.84 to 7.00 times) for all the datasets except the Car dataset, 
which indicated that our proposed method was more suitable for online DA.
Since the Car dataset consists of $d=4096$ dimensional features, 
it consumed more time to compute the mean-target subspace 
as well as the geodesic curve from the source subspace to the mean-target subspace.

\section{Conclusions}
We have described a multi-step framework for tackling the OUDA problem
for classification problem when target data are arriving in mini-batches.
Inspired by the geometrical interpretation of 
computing mean point on the Euclidean space, 
we proposed computing the mean-target subspace 
on the Grassmann manifold incrementally for mini-batches of target data.
We further adopted a feedback step that 
leverages the transformation of the target data at the next timestep.
The transformation matrix 
computed from the source subspace and the mean-target subspace 
aligned the target data closer to the source domain.
Recursive feedback of domain adaptation 
increases the robustness of the recognition system 
for abrupt change of target data.
Fast computation time due to the usage of low-dimensional space 
enables our proposed method to be applied to OUDA in real-time.

\newpage
\bibliographystyle{unsrt}  
\bibliography{paper}


\end{document}